\documentclass{article} 
\usepackage{iclr2024_conference,times}
\usepackage{booktabs}
\usepackage{multirow}

\makeatletter
\renewcommand{\@maketitle}{\vbox{\hsize\textwidth
{\LARGE\sc \@title\par}
\vspace*{1.5cm}

\lhead{}
\def\And{\end{tabular}\hfil\linebreak[0]\hfil
        \begin{tabular}[t]{c}\rule{\z@}{24pt}\ignorespaces}%
\def\AND{\end{tabular}\hfil\linebreak[4]\hfil
        \begin{tabular}[t]{c}\rule{\z@}{24pt}\ignorespaces}
\begin{center}
\begin{tabular}[t]{c}\rule{\z@}{24pt}\@author\end{tabular}
\end{center}
\vskip 0.3in minus 0.1in}}
\makeatother

\usepackage{amsmath,amsfonts,bm}

\def\eqref#1{equation~\ref{#1}}

\def\1{\bm{1}}

\DeclareMathAlphabet{\mathsfit}{\encodingdefault}{\sfdefault}{m}{sl}
\SetMathAlphabet{\mathsfit}{bold}{\encodingdefault}{\sfdefault}{bx}{n}

\usepackage{hyperref}
\usepackage{url}

\usepackage[utf8]{inputenc} 
\usepackage[T1]{fontenc}    
\usepackage{hyperref}       
\usepackage{url}            
\usepackage{booktabs}       
\usepackage{amsfonts}       
\usepackage{nicefrac}       
\usepackage{microtype}      
\usepackage{xcolor}         

\usepackage{dsfont}
\usepackage{amsmath,amssymb,amsthm,mathtools}

\usepackage{wrapfig}
\usepackage{graphicx}
\usepackage{float}
\usepackage{enumitem}
\usepackage{subcaption}

\usepackage{tikz}
\usepackage{listings}       
\usepackage[utf8]{inputenc} 
\usepackage[T1]{fontenc}    
\usepackage{hyperref}       
\usepackage{url}            
\usepackage{booktabs}       
\usepackage{amsfonts}       
\usepackage{nicefrac}       
\usepackage{microtype}      
\usepackage{xcolor}         

\definecolor{codegreen}{rgb}{0,0.6,0}
\definecolor{codegray}{rgb}{0.5,0.5,0.5}
\definecolor{codepurple}{rgb}{0.58,0,0.82}
\definecolor{backcolour}{rgb}{0.95,0.95,0.92}

\lstdefinestyle{mystyle}{
    backgroundcolor=\color{backcolour},   
    commentstyle=\color{codegreen},
    keywordstyle=\color{magenta},
    numberstyle=\tiny\color{codegray},
    stringstyle=\color{codepurple},
    basicstyle=\ttfamily\footnotesize,
    breakatwhitespace=false,         
    breaklines=true,                 
    captionpos=b,                    
    keepspaces=true,                 
    numbers=left,                    
    numbersep=5pt,                  
    showspaces=false,                
    showstringspaces=false,
    showtabs=false,                  
    tabsize=2
}

\usepackage{algpseudocode} 
\usepackage{algorithm}

\usepackage{multirow}

\title{Beyond Tokens in Language Models: Interpreting Activations through Text Genre Chunks}

\author{
    \begin{tabular}{cc}
      \begin{tabular}[t]{c}
        \textbf{Éloïse Benito-Rodriguez}\\
        Independent \\
        \texttt{eloise\_benito@laposte.net}
      \end{tabular}
      &
      \begin{tabular}[t]{c}
        \textbf{Einar Urdshals}\\
        Chalmers Technical University \\
        \texttt{einarurdshals@gmail.com}
      \end{tabular}
      \\
      \addlinespace[2.5ex]
      \begin{tabular}[t]{c}
        \textbf{Jasmina Nasufi}\\
        Lund University \\
        \texttt{jasminanasufi9@gmail.com}
      \end{tabular}
      &
      \begin{tabular}[t]{c}
        \textbf{Nicky Pochinkov} \\
        Independent \\
        \texttt{paper@nicky.pro}
      \end{tabular}
    \end{tabular}
}

\iclrfinalcopy 

\begin{document}

\maketitle

\begin{abstract}
Understanding Large Language Models (LLMs) is key to ensure their safe and beneficial deployment. This task is complicated by the difficulty of interpretability of LLM structures, and the inability to have all their outputs human-evaluated. In this paper, we present the first step towards a predictive framework, where the genre of a text used to prompt an LLM, is predicted based on its activations. Using \texttt{Mistral-7B} and two datasets, we show that genre can be extracted with F1-scores of up to $98\%$ and $71\%$ using \texttt{scikit-learn} classifiers. Across both datasets, results consistently outperform the control task, providing a proof of concept that text genres can be inferred from LLMs with shallow learning models.

\end{abstract}

\section{Introduction}

As language models continue to improve in performance and are applied in a growing number of sectors, it is becoming more important to understand how these models work and monitor their outputs. However, the current paradigm of transformer interpretability focuses almost exclusively on predicting single tokens outputs, with some limited research looking at a specific component of text such as truthfulness or emotion.

We lay the groundwork for looking at language model outputs on the scale of clusters of tokens as interpretable natural language separations. By looking at larger blocks of text, we aim to make it easier to interpret the highest-level of predictions that a language model is making, and to make it easier to monitor large scale language models outputs.

We achieve this with two datasets. The first dataset is manually constructed using a small dataset of diverse language outputs, which we then curate using by machine- and human-labelling of different chunks of these texts. The second is an established dataset, CORE \citep{corpus}. We show that the text chunks in these two datasets are interpretable and easy to classify using simple probes, and that these outputs may be relatively natural units of study.
We provide all code and our dataset on GitHub \footnote{
Code available at \url{https://github.com/Aza-Spearal/Trajectories-Probing}.
}.

\section{Related Work}

Most mechanistic interpretability research focuses on understanding the effects neuron activations have on single token activations \citep{DBLP:conf/emnlp/GevaSBL21, causal-scrubbing, acdc, logit-lens, tunedlens, circuits-induction-heads}. While this is an important and natural bottom-up lens through which to understand models, it stands that a more top-down approach may be underexplored.

Recent work on investigating language model activations and Representation Engineering \citep{representation-engineering, dlk-discovering-latent-knowledge, act-add, language-models-space-and-time, rome-paper}, suggests that it is possible to extract human-understandable concepts with their activations. These results support the hypothesis that a top-down approach to language model interpretability like the method described in this paper is tractable.

In addition, research on modularity and activation sparsity in language models \citep{deja-vu-contextual-sparsity, pochinkov2023dissecting, moefication, emergent-modularity, modular-deep-learning}, as well as the achievement of high pruning ratios for task-specific models, \citep{contrastive-pruning-compression} suggest there may be separable ways of looking at different components of text.

Research on image models, including interpretability research \citep{olah-branch-specialization, olah-feature-vis, olah-differentiable-image-parametrization}, and research into the machine unlearning of specific classes and concepts \citep{selective-synaptic-dampening, machine-unlearning-2021, survey-of-machine-unlearning}, highlights that it is possible to understand how models activate not only on the specific pixel values, but also on larger scale concepts. Our work tries to lay the groundwork for finding these larger scale concepts in text.

There exists a large variety of research into using language models for text embedding \citep{mteb-text-embedding} and classification \citep{text-classification-metareview, universal-text-classification} on various datasets and tasks, including
news categorisation \citep{text-classification-datasets}, 
sentiment analysis \citep{imdb-sentiment-analysis-dataset}, 
question answering \citep{question-answering-classification}, semantic relatedness \citep{classification-semantic-relatedness}, 
and retrieval augmentation \citep{beir-text-embedding-retrieval, retrieval-ranking}. However, these are focused on using language models to perform specific tasks, rather than understanding how a language model performs tasks.

Relatedly, there is research on topic analysis of single texts \citep{thematic-analysis-chatgpt} and large corpuses of text \citep{computational-thematic-analysis-toolkit},
as well as research into qualitative deductive coding, \citep{qualitative-coding-qualigpt, qualitative-coding-patat, qualitative-coding-cody}.
The research is often narrow in scope, such as finding aggregate themes of outputs from specific online communities 
\citep{textual-analysis-chatgpt, deductive-coding-content-analysis-chatgpt}, 
 classifying types of questions 
 \citep{qualitative-coding-kids-questions-gpt3}, or extracting and reformatting knowledge \citep{knowledge-graphs-chatgpt}. 
 Our research, instead of topics, aims to probe into the working of language models to understand on a high-level, how different multi-token scale components are represented in the activations, and thus investigate text categories \citep{helpful-professor-18-text-types} and block components.

\section{Method}

We define chunks as natural units of text that are split into different categories. For the most part, these chunks are analogous to paragraphs, and are simply split by double newlines \texttt{\textbackslash n\textbackslash n}, though with some few exceptions. For details, see Appendix \ref{app:examples}. 

The goal of this work is to investigate the extent to which different classifiers manage to predict the category of a text chunk, given the activations and residual stream that this text generates in a pre-trained model. To this end, we create a dataset and train various classifiers as prediction models. We describe the process and our methods in the following subsections.

\subsection{Datasets}

\begin{figure}[ht]
    \centering
    \includegraphics[width=0.95\linewidth]{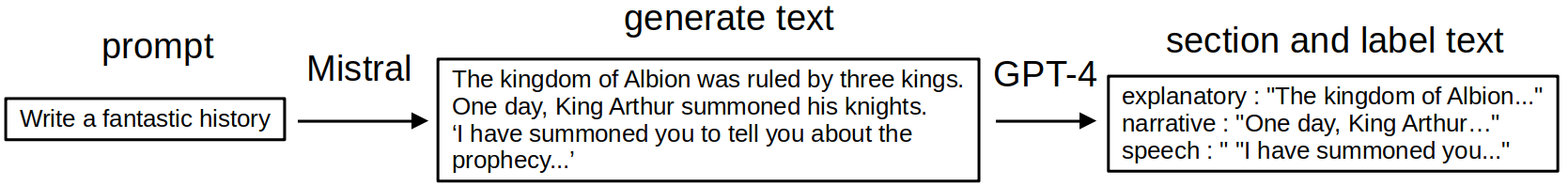}
    \caption{Generation of the labelled dataset.}
    \label{fig:data-creation}
\end{figure}

As shown in Figure \ref{fig:data-creation}, we build our dataset by creating a list of 669 prompts, which are constructed such that they can elicit varied texts from a list of categories and topics. We use them with \texttt{Mistral-7B-Instruct-v2.0} \citep{Mistral} to generate texts. These generated texts are then passed on to \texttt{GPT-4 Turbo Preview} \citep{openai-gpt4}, which is asked to segment the text into chunks according to categories and subsequently label each chunk with the selected category. We obtain 3914 distinct text chunks. The categories we consider are: instructional, explanatory, speech, narrative, and code. The labelling is reviewed by humans and mislabelled text sections are removed. The prompts and examples of the labelled text are shown in the appendix.

The category distribution of this synthetic dataset is shown in Table \ref{tab:data_gen_dist}

\begin{table}[h]
    \centering
    \caption{Distribution of different categories of text chunks.}
\begin{tabular}[h]{@{}lrrrrr@{}}
\toprule
\textbf{Category} & instructional & explanatory & speech & narrative & code \\
\midrule
\textbf{Count} & 1159 & 699 & 548 & 542 & 290 \\
\bottomrule
\end{tabular}
    \label{tab:data_gen_dist}
\end{table}

The primary motivation for using an open-weights model like \texttt{Mistral-7B-Instruct-v2.0} to generate the texts, is that it allows us to access its activations. This lays the groundwork for future work where we predict the sequence of labels based on the activations produced by the prompt.

We additionally chose a second, pre-existing classified dataset to compare our dataset agains.
The chosen dataset was `\textit{Corpus of Online Registers of English}' (CORE) \citep{corpus}. From the dataset, we took the overly fine-grained categories and combined them into fewer larger categories, which we used as the single labels for the entire texts. The category distribution of the CORE dataset is shown in Table \ref{tab:core_cat_dist}

\begin{table}[h]
    \centering
    \caption{Category distribution of the CORE dataset. Top shows category, bottom shows count.}
\begin{tabular}[h]{@{}cccccccc@{}}
\toprule
News & \multirow{2}{*}{Informational} & \multirow{2}{*}{Opinion} & Sports & Personal & \multirow{2}{*}{Persuasion} & \multirow{2}{*}{Discussion} & \multirow{2}{*}{Instructional} \\
Report &  &  & Report & Blog &  &  &  \\
\midrule
6992 & 5917 & 4594 & 1975 & 1849 & 950 & 832 & 214 \\
\bottomrule
\end{tabular}
    \label{tab:core_cat_dist}
\end{table}

\subsection{Training procedure}

We work with the Large language model \texttt{Mistral-7B} \citep{Mistral} with 7-billion parameters. The model is accessed via the Hugging Face Transformers library \citep{DBLP:journals/corr/abs-2211-09085} and run using PyTorch \citep{pytorch}. We prompt the LLM using text chunks from our two datasets, and we study the residual stream and activations at the output of each attention and MLP layer, as described in \citep{gpt2}. Specifically, a text chunk produces activations $a_i^j$ in layer $j$ of the LLM for chunk $c_i$.

\begin{figure}[h]
    \centering
    \includegraphics[width=0.7\linewidth]{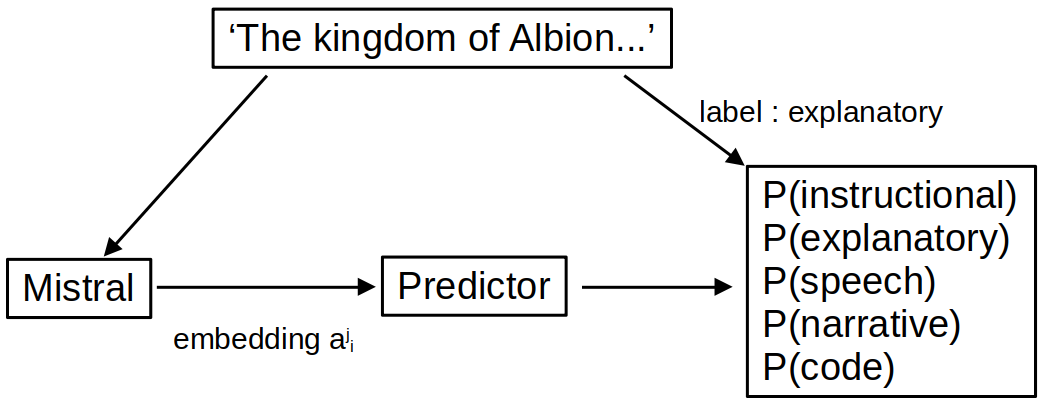}
    \caption{This figure illustrates the training procedure of prediction models on the task of predicting the category of a text section. The activation $a_i^j$ of chunk $c_i$ in the $j$-th layer is extracted from \texttt{Mistral-7B}.}
    \label{fig:training_procedure}
\end{figure}

The means of activations $a_i^j$ serve as input for the probe models we train to predict the correct category for each labelled text chunk. We employ a range of commonly used shallow learning classifiers as our probe models from the scikit-learn library \cite{scikit-learn}. We call them using the Lazy Predict library \cite{lazy_predict}, which enables us to test several scikit-learn algorithms on the same dataset. We keep the vanilla hyperparameters, apart from $\texttt{max\_iter}$ which was fixed at 100000.
We scale our data by removing the mean and scaling to unit variance with $\texttt{StandardScaler()}$.

Probing activations is notoriously subject to all kinds of spurious reasons for high classification accuracy \citep{probing_accuracy}. It demands heavy use of controls in order to validate the study.
We drew inspiration from \citep{probing_control} to provide baselines: for each probe, we built a probe trained and tested on \texttt{Mistral-7B} activations and residual stream, but where all the parameters are random. We use the random model probe results as a control task. If the probes on the original \texttt{Mistral-7B} show better performance than the control task, this will mean that our probes do not rely on spurious correlation, but on true representations in activations.

We divide our dataset into 80\% for the train set and 20\% for the test set.

To evaluate our predictors, we use the Macro F1 score metric \citep{metric} to give the same weight to all classes, even in an unbalanced dataset, including both recall and precision.

\subsection{Dimensional Reduction}

To study how the model might represent different aspects of text, we use a text embedding model, \texttt{Qwen3-Embedding-0.6B} \citep{qwen3embedding} to encode the split chunks of text. For dimensionality reduction and analysis, we use PHATE \citep{phate-dimensionality-reduction}, and show the labeled category. We randomly select 200 samples from each category so as not to overload the display and make it readable.

\section{Results}

\subsection{Dimensional Reduction}
\label{sec:PHATE}

\begin{figure}[h]
    \centering
    \includegraphics[scale=0.25]{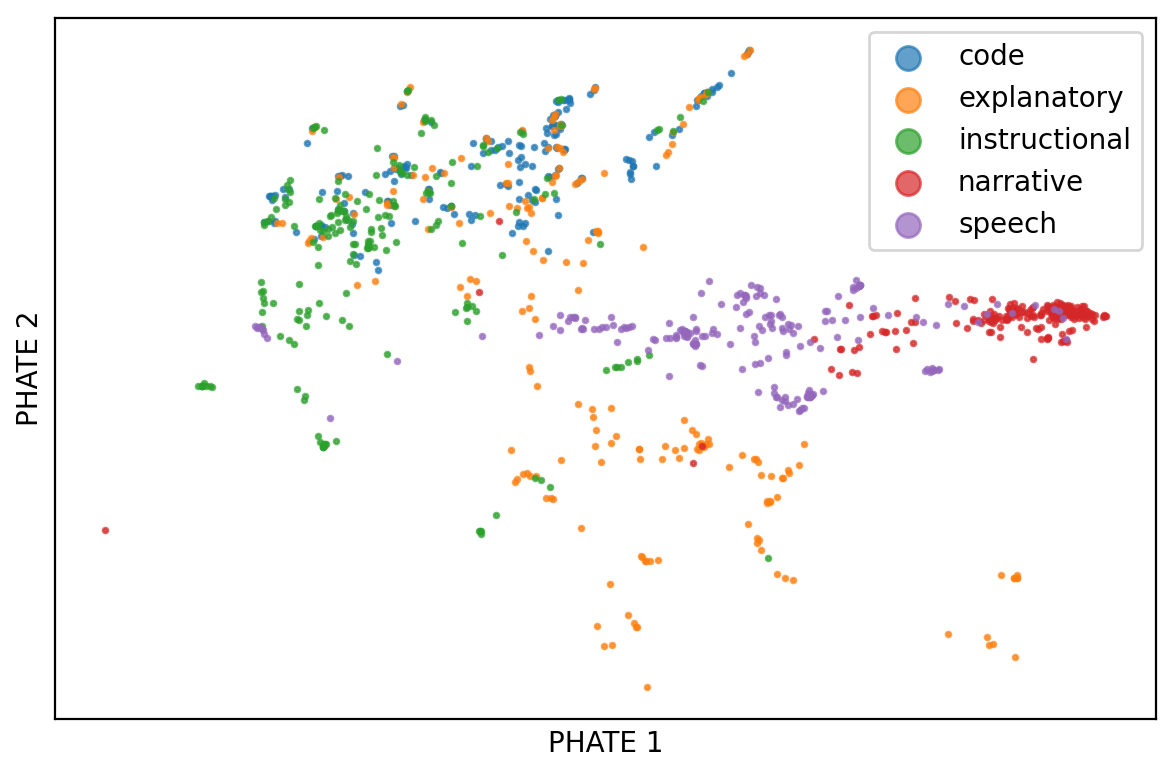}
    \caption{The PHATE dimensionality reduction for the synthetic dataset. We observe that there is some correspondence between the clusters and the labeled categories.}
    \label{fig:phate_human}
\end{figure}

\begin{figure}[h]
    \centering
    \includegraphics[scale=0.69]{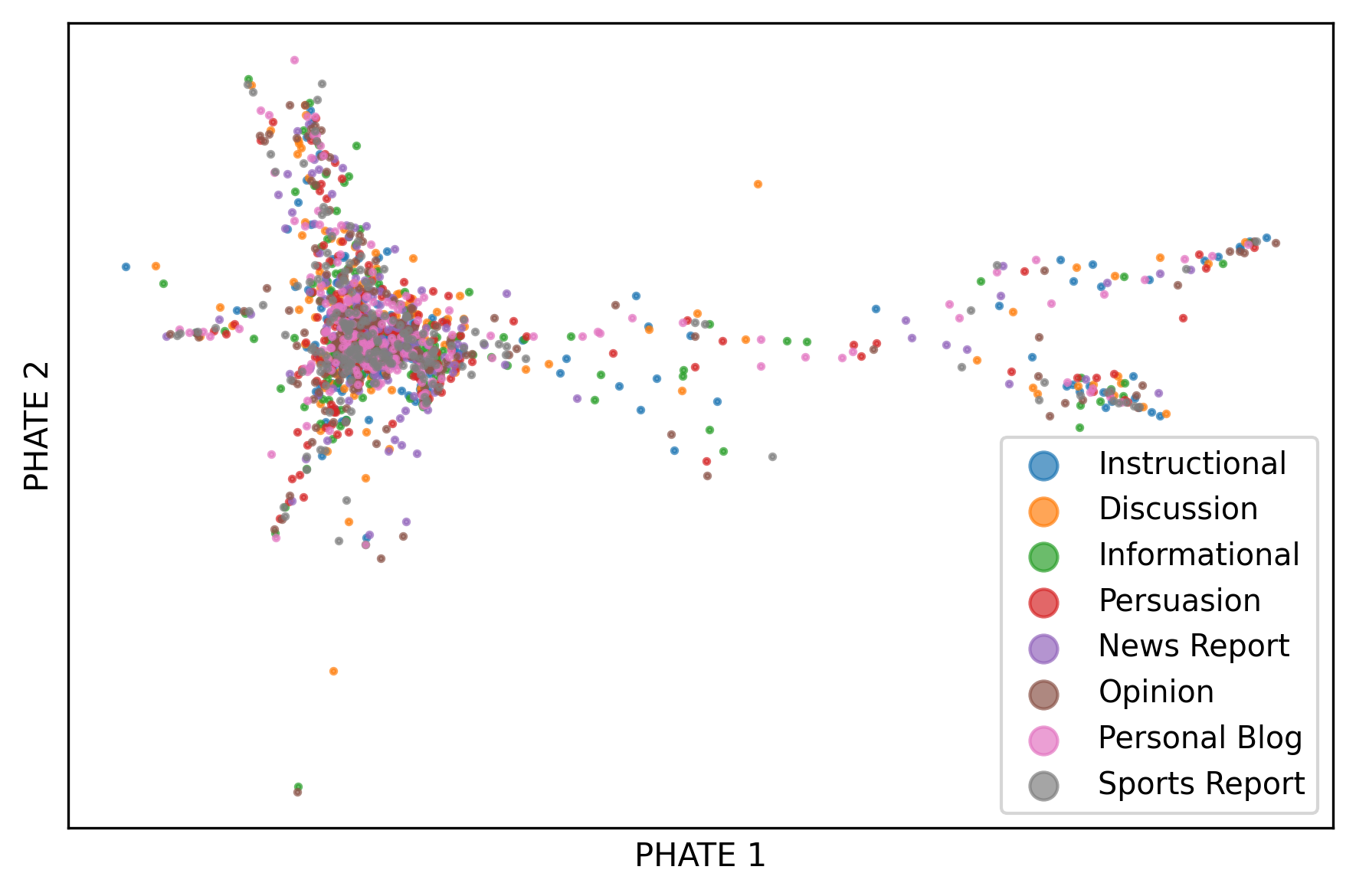}
    \caption{The PHATE dimensionality reduction for the CORE dataset. We see that there is a lot of overlap between clusters and the labelled categories.}
    \label{fig:phate_corpus}
\end{figure}

Looking at Figure \ref{fig:phate_human}, the category does not fully explain the clusters observed in dimensionality reduction, but we see the relatively high correspondence between text categories and various clusters in the embedding space as a promising signal that the encoding of these themes is being captured within the models.

Looking at Figure \ref{fig:phate_corpus}, we see a great overlap with clusters that are difficult to distinguish. This could be explained because there are more classes in the corpus dataset, but also because the categories are much more similar in this dataset.

\subsection{Labelling Accuracy}

\begin{figure}[h]
    \centering
    \includegraphics[scale=0.3]{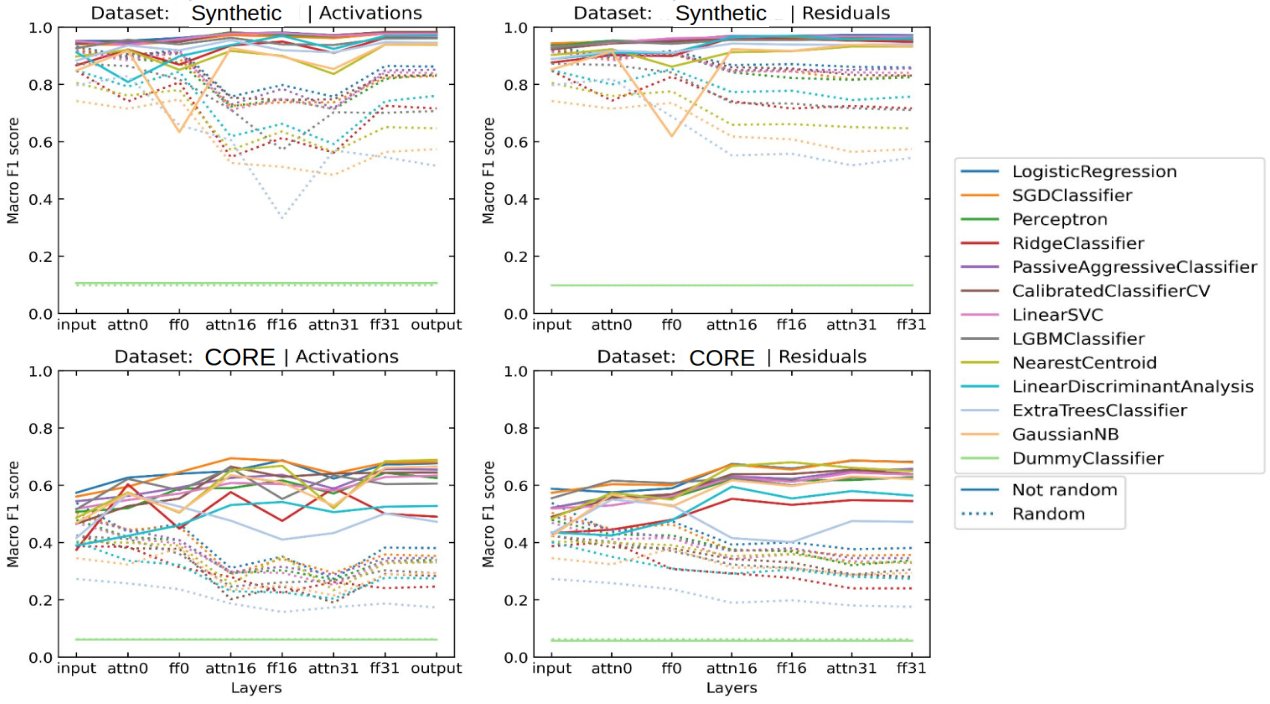}
    \caption{F1-score performance as a function of the layer fraction for prediction models on the task of predicting the category of a text section. The activation $a_i^j$ of chunk $c_i$ at the $j$-th layer has been extracted from \texttt{Mistral-7B}.}
    \label{fig:probing}
\end{figure}

In Figure~\ref{fig:probing} we see the macro F1 score performance for various probe models predicting the category.
The x-axis indicates the layer from which the activations have been extracted, and the colors indicate the predictor that has been employed.
The plain lines correspond to the probing with the test task, whereas the dotted lines correspond to the control task.

The left graphic shows the performance of probe models based on the activations, the right graphic shows the performance based on the residual stream. The top graphic is based on the synthetic dataset, the bottom graphic is based on the CORE dataset.

We see that the deeper we go in the model, the higher the performances.

We see that when the models are trained with the synthetic dataset, the results are largely better than with the CORE dataset, reaching an F1 score of 0.98 whereas the best models built on the CORE dataset reach 0.71. This can be explained by the number of classes which is more important in the CORE dataset. But as we saw in the \hyperref[sec:PHATE]{dimensional reduction results}, the label clusters with PHATE dimensional reduction are reasonable with the synthetic dataset, whereas they are almost indistinguishable with the CORE dataset. We suspect this could mean that text categories are similar in the CORE dataset. Our probe results could be a confirmation of this hypothesis. But contrary to the dimensional reduction, the probes distinguish the labels to a certain extent (better than the random task).

\section{Discussion and Conclusions}
Overall, we find the results promising, and they preliminarily support the hypothesis that models represent high-level multi-token text structures in relatively easy to identify ways. Starting out with 2 datasets, we found that we can achieve a high F1 score for shallow linear probe models, better than the control task. Additionally, our embedding analysis shows interesting results for how models represent concepts.

We saw that the performance of our probe method is superior to the control task, revealing the presence of the representation that we are searching for in the attention patterns. The deeper we go in the Transformer, the more accurate the representations are. The performance is highly dependent on the dataset; we expect that the more different the classes are from each other, the higher the performance will be.

\subsection{Limitations}

We study the activations of \texttt{Mistral-7B}, but we could also probe the activations of other models to see if the results are consistent.
Additionally, we use only two datasets, leaving room to investigate additional datasets to determine what exactly drives the performance.

\subsection{Future Work and Broader Impacts}

Our results indicate that even text categories of moderate overlap, we are able to probe the model relatively accurately, and these methods act as a starting point for probing models beyond the single-token scale.

Future work could investigate scaling up the number of text categories, possibly to include text topic of discussion or other properties of text, and to find better ways of incorporating the overlap of multiple text categories.

Additionally, further work could also try to use these findings and probes to find longer timescale predictions on which categories are likely to emerge in the future of the residual stream, improving our ability to verify the trustworthiness of models stating their intended future actions.

\bibliography{references}
\bibliographystyle{iclr2024_conference}

\newpage
\appendix

\section{Dataset Generation}\label{app:bi-modal}

Our categories are defined in the following way:

narrative: a spoken or written account of connected events; a story. \newline
explanatory: a statement or account that makes something clear. \newline
instructional: detailed information about how something should be done or operated; a direction or order. \newline
speech: spoken words. \newline

We generate our dataset by following these steps:

1. Manually create a list with prompts that can elicit varied texts from an LLM.
The goal is that each element of the list can prompt an LLM to generate text pertaining to particular genres. Additionally, the prompts should be formulated in an open-ended way. The categories we consider are: \{ instructional, speech, narrative, explanatory, code \}. We write 68 prompts that way.\\

2. To have a more diversified dataset, we create synthetic prompts with 
\texttt{Mistral-7B-Instruct-v2.0}
\citep{Mistral} by using the following prompt:

\quad \texttt{``Please generate a prompt list inspired by the list below.'' }
    
\quad \texttt{list\_to\_string(prompts)} 

We remove the prompts that are gibberish or off-topic.

3. We prompt \texttt{Mistral} with our list of prompts to generate the texts with a limit of 500 tokens by output.
    
4. Ask \texttt{gpt-4-turbo-preview} \citep{openai-gpt4} to section and label each text from step 3 according to the following genre categories. We use the following prompt to achieve this:

 \texttt{
 ``Please return a json list that sections the text below and labels it according to one of these categories: instructional, narrative, explanatory, speech, code, other. Please escape characters such as "\textbackslash n". Here is how you should format the output: [\textbackslash n \{"text": ..., "category": ...\}, \textbackslash n \{"text": ..., "category": ...\}, \textbackslash n ... \textbackslash n ]''\\
    (text\_sample)
}

The labelling is reviewed by humans and mislabelled text sections are removed.\\

You can find some examples of how the sectioning and labelling looks like in section \ref{app:examples}.

The genres distribution of this dataset is as follows: `instructional': 1159, `explanatory': 699, `speech': 548, `narrative': 542, `code': 290.

\subsection{Examples}\label{app:examples}

\subsubsection{Example 1: Narrative}
{\bf Prompt:} Create a tale about a person who creates a machine that can predict the future.

\begin{itemize}
    \item \textbf{"text"}: "Title: The Oracular Apparatus", \textbf{"category"}: other

    \item \textbf{"text"}: "Once upon a time, in the heart of the bustling city of Zephyra, resided an unassuming man named Archimedes. He was a renowned inventor, scientist, and philosopher, known for his relentless curiosity and insatiable thirst for knowledge. However, Archimedes had a secret ambition, a dream that would change the course of history - the invention of a machine that could predict the future.", \textbf{"category"}: narrative
    
    \item \textbf{"text"}: "Archimedes toiled away in his laboratory, day and night, fueled by his determination and driven by his relentless pursuit. He contemplated the concept of time, the tangible yet elusive force that governed all existence. He pondered over the intricacies of the universe and the delicate balance of cause and effect. His mind was a whirlwind of thoughts, and he knew that if he could unlock the secrets of time, he could create a machine that could see into the future.", \textbf{"category"}: narrative

    \item \textbf{"text"}: "Driven by his obsession, Archimedes began constructing a device that he came to call the Oracular Apparatus. The machine was a marvel, a majestic contraption that resembled a vast, mechanical cosmos. It was adorned with gears and cogs, pendulums and pulleys, all intricately connected to form a harmonious, self-contained entity. The apparatus was designed to observe the patterns and connections in the universe and to decipher the inherent formulae that governed the passage of time.", \textbf{"category"}: explanatory

    \item \textbf{"text"}: "As Archimedes fine-tuned his machine, he discovered that it was capable of depicting the future in the form of visions, images that revealed the events that were yet to unfold. Mesmerized by his creation, Archimedes spent hours gazing into the Oracular Apparatus, fascinated by the glimpses of the future that unfurled before his eyes.", \textbf{"category"}: narrative

    \item \textbf{"text"}: "News of Archimedes' Oracular Apparatus spread like wildfire. People from all corners of Zephyra flocked to the inventor's laboratory, seeking answers to their most pressing questions and desperate for a peek into the future. The machine's accuracy was unparalleled, and its prophecies became the talk of the town.", \textbf{"category"}: narrative

     \item \textbf{"text"}: "However, as time went by,", \textbf{"category"}: other
     
\end{itemize}

\subsection{Example 2: Coding}
{\bf Prompt:} "Implement a class in C++ representing a geometric shape, with methods to calculate its area and perimeter.
\begin{itemize}
    \item \textbf{"text"}: "Here's an example of a simple Shape base class and a Derived class Circle that follows your requirements:", \textbf{"category"}: other

    \item \textbf{"text"}:
    \begin{minipage}[t]{\linewidth}
     \begin{lstlisting}
     // C++ program with Shape and Circle classes
     #include <iostream>
     #include <cmath>
     
     // Base Shape class
     class Shape {
     public:
         virtual double area() const = 0;
         virtual double perimeter() const = 0;
     };
     
     // Circle class derived from Shape
     class Circle : public Shape {
     private:
         double radius;
     
     public:
         Circle(double r) : radius(r) {}
     
         double area() const override {
             return M_PI * pow(radius, 2);
         }
     
         double perimeter() const override {
             return 2 * M_PI * radius;
          }
     };
     
     // Test function
     int main() {
         Circle circle(5);
         std::cout << "Area: " << circle.area() << std::endl;
         std::cout << "Perimeter: " << circle.perimeter() << std::endl;
     
         return 0;
     }
     \end{lstlisting}
    \end{minipage}
    , \textbf{"category"}: code

    \item \textbf{"text"}: "In this example, we have a base `Shape` class with pure virtual functions `area()` and `perimeter()`. The `Circle` class is derived from `Shape`, and it provides the implementation for both methods. In the provided `main` function, we create a `Circle` object and calculate its area and perimeter.", \textbf{"category"}: explanatory
       
\end{itemize}

\end{document}